%% file: main_v3_camera_ready.tex
\documentclass[10pt,letterpaper]{article}
\usepackage{cogsci}
\cogscifinalcopy
\input{config}

\title{Timing is Everything: Temporal Scaffolding of Semantic Surprise in Humor}

\author{%
    Yuxi Ma$^{1,2,5,6\,*}$, Yongqian Peng$^{1,2,4,5,6\,*}$, Junchen Lyu$^{1,4}$, Chi Zhang$^{3,5\,\textrm{\Letter}}$, and Yixin Zhu$^{2,1,5,6\,\textrm{\Letter}}$
    \vspace{3pt}\\\normalfont    
    \small $^\star{}$Equal contributors\quad{}
    \small Project Website: \url{https://mayuxi.com/research/talkshow}\\
    \small $^1$ Institute for Artificial Intelligence, Peking University\quad
    \small $^2$ School of Psychological and Cognitive Sciences, Peking University\\
    \small $^3$ School of Intelligence Science and Technology, Peking University\quad
    \small $^4$ Yuanpei College, Peking University\\
    \small $^5$ State Key Laboratory of General Artificial Intelligence, Peking University\\
    \small $^6$ Beijing Key Laboratory of Behavior and Mental Health, Peking University
    \vspace{-12pt}
}

\begin{document}
\begin{CJK*}{UTF8}{gbsn}

\maketitle

\begin{abstract}
Humor is a fundamental cognitive phenomenon in which humans derive pleasure from the expectation violations and their resolution, exemplifying the brain’s dynamic capacity for predictive processing.
Classical humor theories emphasize semantic incongruity as the primary driver of amusement, yet overlook temporal dynamics despite comedians' intuition that ``timing is everything.'' The extent to which temporal structure contributes to humor appreciation and how it interacts with semantic content remains poorly understood.
Here, we propose the \ac{dpv} framework to capture the interplay between content and timing. By analyzing 828 professional Chinese stand-up performances, we show that temporal features substantially outweigh semantic incongruity in predicting audience appreciation.
Specifically, we find that peak semantic violations matter more than average incongruity levels, and pauses systematically lengthen before high-surprise punchlines---a strategic coupling that distinguishes successful from unsuccessful performances.
These findings reframe humor as temporally scaffolded, where timing and semantic content operate in strategic coordination rather than independently.
Our \ac{dpv} framework bridges humor theory with predictive processing, demonstrating that temporal structure plays a central role in naturalistic humor appreciation with implications for understanding multi-scale prediction integration in linguistic processing.

\end{abstract}

\begin{figure*}[ht!]
    \centering
    \small
    \includegraphics[width=0.8\linewidth]{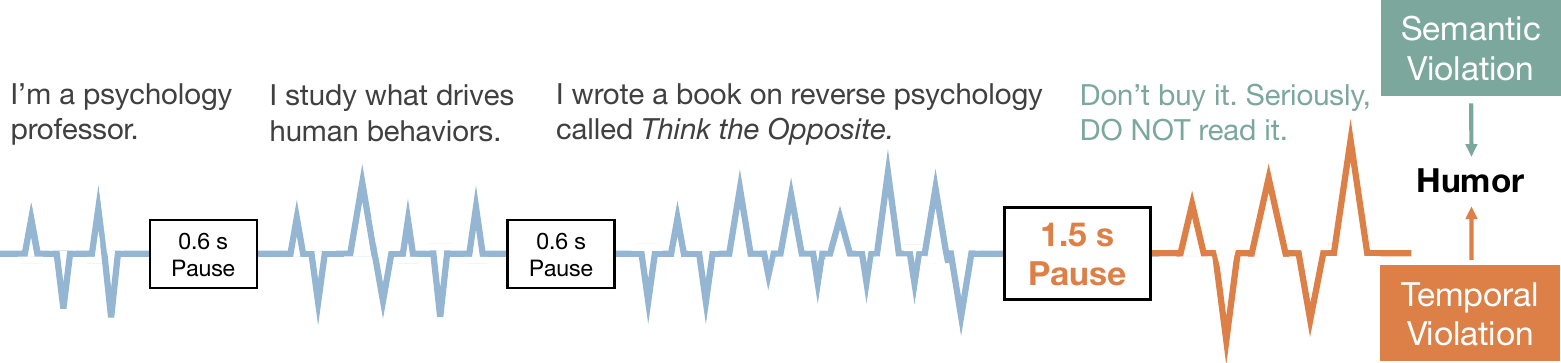}
    \caption{\textbf{Illustration of the proposed \acf{dpv} mechanism.} Setup statements establish temporal and semantic expectations through regular patterns (0.6s pauses, small semantic distances between consecutive sentences). The punchline simultaneously violates both dimensions: an extended pause (1.5s) disrupts established rhythm while semantically distant content deviates sharply from contextual predictions. This strategic coupling of temporal and semantic violations creates conditions for efficient prediction error resolution, generating humor.}
    \label{fig:framework}
\end{figure*}

\section{Introduction}

Humor offers a unique window into how humans manage expectations and derive reward from prediction violations. Classical theories, particularly the two-stage incongruity-resolution model~\citep{suls1972two,suls1983cognitive}, propose that humor arises when audiences detect and resolve semantic violations. Recent developments in the predictive processing framework have refined this view~\citep{van2017affective}, proposing that humor emerges not merely from resolving incongruity but from the \textit{efficiency} of resolution. Critically, \citet{van2017affective} argue that pleasure depends on the \textit{speed} of understanding---the brain rewards itself for unexpectedly rapid error correction, a metacognitive pleasure signal tied to resolution dynamics rather than content alone.

This raises a fundamental question: what determines resolution efficiency? We propose that the answer lies in temporal structure. Existing frameworks focus overwhelmingly on \textit{what} violates expectations (semantic content), while neglecting \textit{when} and \textit{how} these violations unfold. Yet temporal scaffolding---pacing, pauses, rhythmic patterns---may fundamentally shape whether prediction errors resolve swiftly (producing pleasure) or stall (producing confusion). Timing does not merely deliver semantic surprise; it may actively gate cognitive access to that content.

This theoretical gap is partly methodological. As \citet{moran2004neural} noted, prior literature has relied heavily on decontextualized stimuli: written jokes, isolated punchlines, or brief text strings. While such controlled materials successfully isolate semantic incongruity mechanisms, they eliminate the rich temporal dynamics that characterize professional comedic delivery. Consequently, we lack quantitative evidence for how temporal dynamics interact with semantic surprise to drive audience engagement in ecologically valid settings.

The current study addresses this gap by analyzing naturalistic performances at scale. We created a dataset of 828 professional Chinese stand-up comedy performances from major televised competitions (2017--2025), representing 86 hours of performance time with accompanying audience appreciation scores. While the televised format introduces editing decisions, professional editors work to maintain narrative coherence and pacing that align with the audience's cognitive rhythms rather than arbitrarily distorting the temporal structure. This analytical approach allows us to disentangle the relative contributions of content and timing without the constraints of artificial laboratory environments.

We propose a \acf{dpv} framework: humor arises from the interplay between two prediction systems (see \cref{fig:framework}). \textit{Semantic prediction violations} occur when content deviates from contextually expected continuations. \textit{Temporal prediction violations} occur when pacing and pause patterns deviate from established rhythmic expectations, inspired by dynamic attending~\citep{large1999dynamics}. Critically, these systems interact hierarchically: temporal structure acts as a cognitive gatekeeper determining \textit{when} semantic information is processed relative to attentional peaks. Extended pauses before high-surprise content create temporal windows for priming semantics and accumulating prediction errors; when the punchline arrives within this optimal window, pre-activated cognitive resources enable rapid resolution, thereby maximizing the efficiency gain that drives pleasure.

To test this framework, we analyzed the 828 Chinese stand-up comedies, quantifying temporal dynamics through pause patterns and speech rate, alongside semantic incongruity via embedding-based distance measures. Our analyses reveal three key findings. First, temporal features substantially outperform semantic features in predicting audience appreciation, suggesting that timing dynamics provide the primary mechanism for humor success. Second, semantic violations show their strongest effects at peak moments---defined by maximum deviation from contextual expectations rather than average incongruity---consistent with predictive processing's emphasis on salient prediction errors. Third, sentence-level analysis of high-performing sets reveals systematic coordination between timing and content: pauses are extended more strongly before high-surprise content compared to low-performing sets. This differential coupling pattern is consistent with the temporal scaffolding of semantic processing, repositioning timing from a delivery detail to a potential cognitive mechanism in humor appreciation.

\section{Related Work}

\subsection{Humor Theories and Computational Approaches}

Classical incongruity theory posits that humor arises from detecting and resolving semantic violations~\citep{suls1972two,suls1983cognitive}. Jokes succeed when audiences first encounter content violating contextual expectations, then discover alternative interpretations rendering violations meaningful. \citet{attardo1991script} systematized this view in their \ac{gtvh}, proposing six knowledge resources---including script opposition and logical mechanisms---that characterize joke structure. These frameworks dominated humor research for decades, positioning semantic content as the primary determinant of funniness.

Recent theoretical advances have integrated the principles of predictive processing. \citet{van2017affective} argue that pleasure emerges not from mere incongruity resolution, but from \textit{efficient} resolution---when the brain resolves prediction errors faster than anticipated, it generates amusement as a metacognitive reward signal. This framework yields a critical prediction: any factor that modulates resolution speed should also modulate humor intensity. Neuroimaging supports dissociable neural correlates for joke comprehension \vs perceived funniness~\citep{moran2004neural,vrticka2013neural}, confirming that understanding and pleasure reflect distinct processes.

Computational approaches have increasingly operationalized these theoretical constructs. Early work identified semantic violations through wordplay detection, script opposition, or lexical substitution~\citep{cattle2018recognizing,veale2004incongruity}. Contemporary methods leverage distributional semantics to quantify semantic surprise: embedding-based approaches measure punchline deviation from setup context in semantic space, operationalizing incongruity as vector distances in BERT or GPT~\citep{mihalcea2010computational,kao2016computational,annamoradnejad2024colbert}. These computational measures effectively predict humor ratings in written text, demonstrating that quantifiable semantic features capture core aspects of funniness.

Yet these approaches, theoretical and computational alike, share a critical limitation: they treat humor primarily as a \textit{content} problem. Timing appears, if at all, as a delivery variable amplifying inherently funny semantic structures. This reflects methodological constraints: whether analyzing written jokes or laboratory presentations with fixed timing, existing work eliminates the temporal variability characterizing naturalistic performance~\citep{moran2004neural}. More fundamentally, if humor pleasure derives from resolution efficiency as \citet{van2017affective} propose, \textit{temporal dynamics} of information delivery should centrally determine when and how efficiently audiences resolve incongruity. Yet quantitative analyses examining timing-semantic interactions remain sparse.

\subsection{Temporal Dynamics in Cognition}

Research beyond humor demonstrates that temporal structure fundamentally shapes cognitive processing. Dynamic attending theory shows that rhythmic regularities generate anticipatory neural states, modulating when information is processed most efficiently~\citep{large1999dynamics}. Audiences entrain to temporal patterns, creating attentional windows at predicted moments~\citep{jones1989dynamic}. Violations of these temporal expectations (unexpected pauses or accelerations) force attentional reorientation, potentially priming cognitive systems for salient information~\citep{barnes2000expectancy}.

Suspense research provides converging evidence that temporal scaffolding determines semantic processing~\citep{lehne2015toward}. \citet{lehne2015reading} demonstrated that extended anticipatory periods before resolution engage predictive mechanisms, with neural activity reflecting accumulated prediction error. Critically, anticipation \textit{duration} modulates both prediction error intensity and subsequent reward signals upon resolution~\citep{fiorillo2003discrete}. In speech processing, prosodic boundaries marked by pauses signal upcoming information structure, enabling listeners to prepare appropriate cognitive resources~\citep{bogels2011prosodic}.

Despite these insights, temporal dynamics remain underexplored in humor research~\citep{attardo2001humorous}. While some studies acknowledge comic timing matters~\citep{norrick2001conversational}, quantitative analyses examining how pause patterns, speech rate, and rhythmic structure interact with semantic surprise are scarce.

\section{Methods}

\subsection{Dataset}

We compiled an initial corpus of 1102 professional Chinese stand-up comedy performances totaling 107 hours, broadcast between 2017 and 2025. The dataset comprises five seasons of \textit{Rock \& Roast} (\textit{脱口秀大会}), two seasons of \textit{Stand-up Comedy and Friends} (\textit{脱口秀和TA的朋友们}), and two seasons of \textit{Comedy King} (\textit{喜剧之王}). We processed each performance using timestamped \ac{asr} to extract temporal and semantic features.

Audience appreciation was measured through live voting during each performance. Our primary analyses focus on 828 performances (86 hours) with complete voting data. To account for heterogeneous voting systems across shows, we normalized votes by calculating the vote rate (votes received divided by total possible votes). This normalization enables comparison across shows while preserving relative performance quality within each competition. The mean performance duration was 373.06 seconds ($SD = 83.49$ seconds).

\subsection{Feature Extractions}

\subsubsection{Temporal Features}

From timestamped \ac{asr} transcripts, we extracted temporal dynamics capturing baseline pacing and rhythmic variability. \textit{Average pause duration} measures mean silence length between speech segments (seconds), quantifying baseline temporal spacing. \textit{Pause variability} captures the standard deviation of pause durations (seconds), operationalizing temporal unpredictability in rhythmic structure. \textit{Speech rate} is computed as characters (syllables) per second during speech, reflecting delivery speed; in Mandarin Chinese, each written character corresponds to a single spoken syllable. These features capture distinct aspects of temporal structure relevant to audience prediction dynamics: average pause duration and speech rate establish baseline temporal expectations, while pause variability operationalizes deviations from established rhythmic patterns.

\subsubsection{Semantic Features}

We quantified semantic incongruity through embedding-based distances between consecutive sentences~\citep{ma2025word}. Transcripts were automatically segmented into sentences by the \ac{asr} system. For each consecutive sentence pair, we computed embeddings using OpenAI's text-embedding-3-small model and calculated semantic distance as $\textit{1 - \text{cosine similarity}}$ from L2-normalized embeddings. Higher distances indicate greater deviation from preceding context, directly operationalizing sentence-level semantic incongruity.

For each performance, we extracted distributional statistics capturing different aspects of semantic violation. \textit{Average distance} measures overall incongruity across the performance. \textit{Peak distance} captures maximum incongruity, identifying the sentence deviating most strongly from its predecessor. \textit{Distance trend} quantifies whether semantic distances systematically change over time, computed as the slope of distance regressed on sentence position. \textit{Distance shift} measures whether distances are higher in the second versus the first half, capturing structural changes in semantic patterning.

\section{Results}

\subsection{Temporal Dynamics and Audience Appreciation}

We examined relationships between features and audience appreciation using partial correlations controlling for performance duration, as duration naturally correlates with many structural aspects. Group comparisons contrasted high-performing (top 20\%, $n=171$) and low-performing (bottom 20\%, $n=169$) sets using independent samples t-tests with Cohen's $d$ effect sizes. The slight group size difference reflects tied vote rates at cutoff boundaries.

\begin{figure}[t!]
    \centering
    \includegraphics[width=0.75\linewidth]{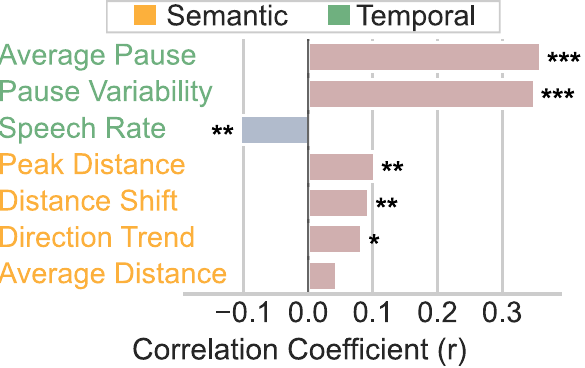}
    \caption{\textbf{Correlation between features and audience appreciation.} Partial correlation coefficients (controlling for performance duration) between temporal features (green labels) and semantic features (yellow labels) with audience vote rates. Temporal dynamics substantially outperform semantic features in predicting appreciation, with average pause duration and pause variability showing the strongest effects. Statistical significance: ${}^{*}p < 0.05$, ${}^{**}p < 0.01$, ${}^{***}p < 0.001$.}
    \label{fig:predictors}
\end{figure}

Temporal dynamics emerged as the strongest predictor of audience voting behavior (see \cref{fig:predictors}), substantially outweighing all other features. Average pause duration demonstrated a strong positive correlation with votes (partial $r = 0.36$, $p < 0.001$), representing one of the largest effects observed. Pause variability (standard deviation of pause durations)---capturing temporal unpredictability---also predicted engagement robustly (partial $r = 0.35$, $p < 0.001$). These findings align with dynamic attending theory \citep{large1999dynamics}, suggesting performers strategically manipulate temporal expectations to heighten anticipation and amplify subsequent surprise.

Speech rate showed a negative correlation with votes (partial $r = -0.10$, $p < 0.01$), indicating slower delivery facilitates engagement. This pattern suggests rushed pacing may undermine comedic impact, potentially by providing insufficient time for prediction generation. Combined with pause effects, this indicates temporal control involves both strategic silence (extended pauses) and modulated speech (slower rate), jointly expanding the temporal window for predictive processing.

\begin{figure}[t!]
    \centering
    \includegraphics[width=\linewidth]{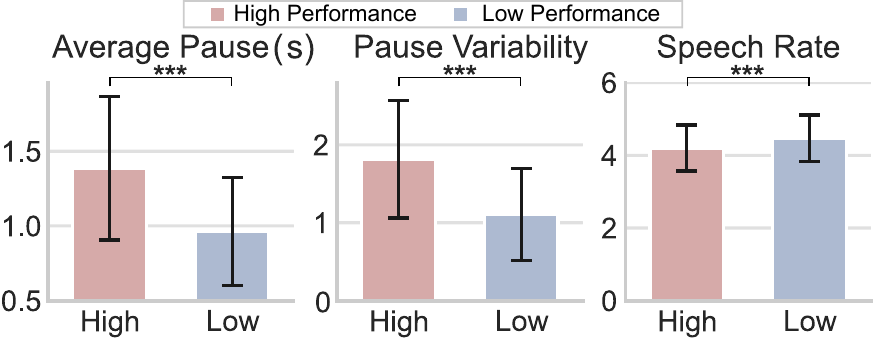}
    \caption{\textbf{Group comparisons of temporal features.} Mean values for high-performing (top 20\%, red) and low-performing (bottom 20\%, blue) sets across three temporal dimensions: average pause duration (seconds), pause variability (seconds), and speech rate (characters per second). High-performing comedians exhibit substantially longer and more variable pauses, alongside slower speech rates. Error bars represent standard deviation. Statistical significance: ${}^{***}p < 0.001$.}
    \label{fig:temporal}
\end{figure}

Group comparisons provided converging evidence with large effect sizes (see \cref{fig:temporal}). High-performing sets exhibited substantially longer average pauses ($M_{high} = 1.39$ s, $SD = 0.48$ s \vs $M_{low} = 0.96$ s, $SD = 0.36$ s; $t(322) = 8.96$, $p < 0.001$, Cohen's $d = 0.99$). Pause variability showed similar discrimination ($M_{high} = 1.81$ s, $SD = 0.75$ \vs $M_{low} = 1.10$ s, $SD = 0.59$; $t(322) = 9.48$, $p < 0.001$, $d = 1.05$). Speech rate differences were smaller but consistent: successful performers spoke more slowly ($M_{high} = 4.20$, $SD = 0.63$ \vs $M_{low} = 4.47$, $SD = 0.64$; $t(322) = -3.72$, $p < 0.001$, $d = -0.41$).

\subsection{Semantic Violations and Audience Appreciation}

Semantic distance measures supported incongruity-based mechanisms, though effect sizes were smaller than those for temporal features (see \cref{fig:predictors}). Peak distance---representing maximum semantic deviation in a performance---correlated with audience appreciation (partial $r = 0.10$, $p < 0.01$). Distance shift---measuring whether distances increased from first to second half---predicted appreciation (partial $r = 0.09$, $p < 0.01$), suggesting successful performers escalate unpredictability over time rather than maintaining constant incongruity. Direction trend---quantifying systematic changes in semantic distance throughout performance---showed a positive correlation (partial $r = 0.08$, $p < 0.05$).

\begin{figure}[t!]
    \centering
    \includegraphics[width=0.83\linewidth]{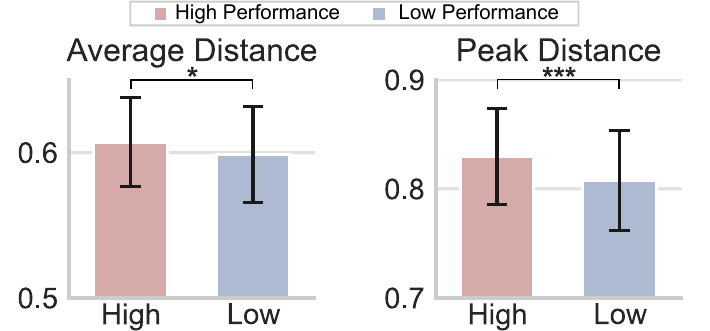}
    \caption{\textbf{Group comparisons of semantic features.} Mean values for high-performing (top 20\%, red) and low-performing (bottom 20\%, blue) sets across semantic dimensions: average distance (left) and peak distance (right). High-performing comedians show greater semantic incongruity, with peak distance demonstrating stronger discrimination than average distance. Error bars represent standard deviation. Statistical significance: ${}^{*}p < 0.05$, ${}^{***}p < 0.001$.}
    \label{fig:semantic}
\end{figure}

Group comparisons mirrored these patterns (see \cref{fig:semantic}). Peak distance was higher in successful performances ($M_{high} = 0.83$, $SD = 0.04$ \vs $M_{low} = 0.81$, $SD = 0.05$; $t(322) = 4.50$, $p < 0.001$, $d = 0.50$). Average distance showed similar but weaker discrimination ($M_{high} = 0.61$, $SD = 0.03$ \vs $M_{low} = 0.60$, $SD = 0.03$; $t(322) = 2.37$, $p = 0.018$, $d = 0.26$).

\subsection{Strategic Coupling: Timing and Content} 

Beyond main effects, we tested whether comedians strategically coordinate timing with content---specifically, whether they amplify pauses before semantically surprising sentences. This \textit{strategic coupling hypothesis} proposes that temporal control scaffolds semantic violations: longer pauses preceding surprising content allow greater prediction generation, amplifying subsequent prediction errors.

\paragraph{Within-performance timing-content coordination}

We analyzed sentence-level relationships between semantic distance and pause duration within individual performances. To isolate strategic timing-content coupling, we defined surprise levels relative to each specific performance. We identified consecutive sentence pairs and classified them based on semantic distance: the top 20\% as high-surprise pairs and the bottom 20\% as low-surprise pairs. This within-performance approach controls for individual stylistic differences; because comedians differ in baseline semantic complexity, global classification would conflate content style with temporal strategy. By comparing high-surprise and low-surprise sentence pairs within a comedian's own set, we tested whether performers systematically modulate timing when introducing semantic violations relative to their established patterns.

\begin{figure}[!b]
    \centering
    \begin{subfigure}[t]{0.49\linewidth}
        \centering
        \includegraphics[width=0.9\linewidth]{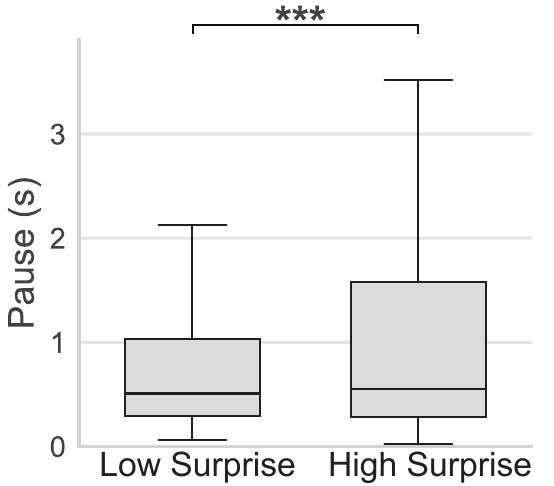}
        \caption{Pause durations across semantic surprise levels}
        \label{fig:pause_surprise}
    \end{subfigure}%
    \hfill
    \begin{subfigure}[t]{0.49\linewidth}
        \centering
        \includegraphics[width=0.9\linewidth]{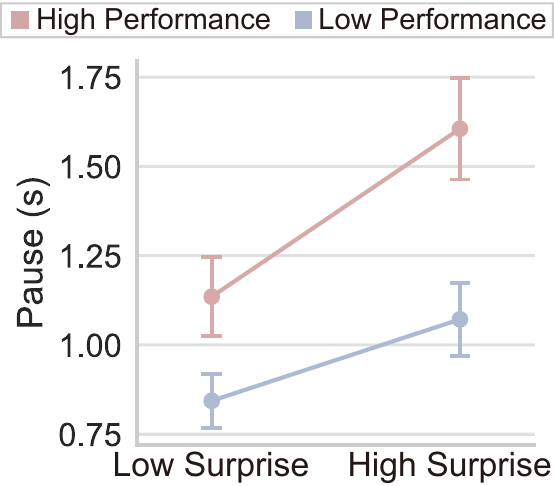}
        \caption{Performance levels moderate timing-content coupling}
        \label{fig:interaction}
    \end{subfigure}%
    \caption{\textbf{Strategic coupling between timing and semantic surprise.} Pauses systematically extend before high-surprise content, with successful comedians showing stronger modulation. (a) Across all performances, pauses before high-surprise pairs (top 20\% semantic distance) are 35.6\% longer than before low-surprise pairs (bottom 20\% semantic distance). (b) High-performing comedians (top 20\%, red) exhibit steeper timing modulation (41.2\% increase) than low-performing comedians (bottom 20\%, blue; 27.4\% increase). Error bars represent the standard error of the mean. Statistical significance: ${}^{***}p < 0.001$.}
\end{figure}

Across all performances, pauses before high-surprise pairs were significantly longer (see \cref{fig:pause_surprise}; $M_{high} = 1.37$s) than those preceding low-surprise pairs ($M_{low} = 1.01$s; $t(1638) = 6.15$, $p < 0.001$). This 0.36-second difference---a 35.6\% increase---demonstrates systematic timing-content coupling: pauses extend systematically before semantically surprising content rather than distributing uniformly across transitions.

\paragraph{Performance levels moderates timing-content coupling}

To determine whether timing-content coordination varies with performance quality, we tested whether successful comedians exhibit stronger coupling than less successful performers.

A two-way ANOVA revealed significant main effects of both performance level ($F(1, 3274) = 49.05$, $p < 0.001$) and surprise level ($F(1, 3274) = 38.35$, $p < 0.001$), plus a significant interaction ($F(1, 3274) = 4.23$, $p = 0.040$). High-performing comedians demonstrated steeper timing modulation (see \cref{fig:interaction}): mean pauses increased from $1.14$s (low-surprise) to $1.61$s (high-surprise), a 41.2\% increase. Low-performing comedians showed weaker modulation, with pauses increasing from $0.84$s to $1.07$s, only a 27.4\% increase. This interaction demonstrates that strategic timing-content coupling is not merely a stylistic choice but may be a marker of comedic expertise. Successful comedians show \textbf{greater differentiation} in pause patterns across semantic surprise levels.

\section{Discussions}

We proposed a \acf{dpv} framework that examines how temporal and semantic mechanisms jointly drive humor appreciation in naturalistic comedy. Three findings emerged: temporal dynamics substantially outperform semantic features in predicting audience engagement; semantic violations have the strongest effects at peak moments; and successful comedians systematically extend pauses before semantically surprising content. Together, these patterns suggest that temporal structure plays a more central role in humor processing than current theories acknowledge.

\subsection{Temporal Dynamics as a Dominant Predictor}

The \ac{dpv} framework hypothesizes that humor arises from coordinated violations of temporal and semantic predictions. Our first finding, that temporal features substantially outperform semantic features, provides initial support for hierarchical organization where temporal dynamics dominate engagement. The average pause duration and the pause variability both showed large effect sizes in group comparisons, while the semantic distance measures showed medium effects. This significant difference in predictive power challenges the predominant focus on semantic content in humor theory~\citep{suls1972two,attardo1991script}. These findings do not imply that semantic content is unimportant---peak semantic distance showed consistent positive correlations with appreciation. Rather, they suggest that among measured factors, temporal structure accounts for substantially more variance in audience responses than semantic incongruity. This asymmetry raises theoretical questions about the relative contributions of ``what'' versus ``when'' in humor processing.

Several mechanisms may explain the dominance of timing. First, pauses create temporal windows for prediction generation. Extended silence allows audiences to elaborate contextual expectations before encountering violations~\citep{lehne2015toward}. The negative correlation with speech rate provides converging evidence: slower delivery extends processing time, potentially facilitating prediction elaboration. Second, pause variability prevents audiences from settling into stable temporal expectations, maintaining cognitive flexibility through dynamic attention~\citep{large1999dynamics}. Third, temporal unpredictability may independently generate prediction errors that prime audiences to detect novelty, heightening sensitivity to subsequent content surprises.

However, our correlational design cannot determine whether the temporal structure directly causes appreciation or reflects other unmeasured factors. High-performing comedians may differ in confidence, stage presence, or audience rapport---qualities that manifest behaviorally as longer pauses and slower delivery.

\subsection{Semantic Violations as Peak Events}

The \ac{dpv} framework predicts that semantic violations should matter, and our findings refine this prediction. Among semantic features, peak distance (maximum semantic deviation) emerged as the strongest predictor of appreciation. This means what matters most is not overall incongruity levels but the presence of particularly salient violations. The average semantic distance showed weaker effects, despite capturing more information about performance-wide incongruity, indicating that audiences respond more to discrete high-surprise moments than to chronic moderate incongruity. This finding aligns with predictive processing accounts that emphasize salient prediction errors~\citep{clark2018nice}. The brain prioritizes resolving large, discrete violations over processing continuous low-level uncertainty. In comedy, this suggests that a few well-placed, surprising moments may be more effective than maintaining a constant, moderate level of incongruity throughout.

The positive correlations of distance shift and direction trend with appreciation provide additional evidence for content structuring strategies. Successful performers appear to escalate semantic violations over time rather than maintain constant incongruity levels. This may create a trajectory of increasing prediction error that sustains engagement, though future research should examine whether this reflects deliberate strategy or emergent properties of joke sequencing.

\subsection{Strategic Coupling Distinguishes Expertise}

The most direct evidence for coordination between temporal and semantic systems came from sentence-level analyses. Across all performances, pauses were 35.6\% longer before high-surprise content compared to low-surprise transitions. Critically, high-performing comedians showed stronger timing-content coupling (41.2\% pause increase) than low-performing comedians (27.4\% increase).

This pattern demonstrates timing-content coordination may be a marker of comedic expertise, but the causal mechanism remains ambiguous. Three interpretations are plausible. First, expert comedians may deliberately extend pauses before surprising content to scaffold audience processing---creating temporal windows for prediction elaboration before violations arrive. Second, semantically complex or surprising content may require more cognitive effort to formulate and deliver, naturally eliciting longer pauses from performers. Third, expertise may manifest as sensitivity to both dimensions: skilled comedians create more surprising content and have better temporal control, with coordination emerging from dual competencies rather than causal scaffolding.

Our data cannot distinguish among these possibilities. However, the interaction effect---where high-performing comedians show \textit{stronger} coupling than low-performing comedians---suggests coordination is not merely an artifact of content complexity. If surprising content simply required a longer delivery time, we would expect similar pause extensions across all performers. That experts show greater coordination suggests either deliberate strategic timing or that expertise involves heightened sensitivity to how content complexity affects temporal demands. Future experimental work manipulating timing while holding content constant could distinguish whether temporal structure causally affects processing or reflects performer-side demands.

The coordination finding is consistent with the temporal scaffolding of semantic processing, but alternative explanations remain viable. Longer pauses before surprising content could reflect comedians' own processing needs, audience feedback mechanisms, or emergent properties of how complex content is constructed during performance. Establishing causal direction and mechanism requires controlled experiments beyond naturalistic observation.

\subsection{Limitations and Future Directions}

Several limitations warrant consideration. First, our use of professional performances in competitive settings provides ecological validity but introduces two interpretive constraints. Audience voting provides a behavioral proxy for comedic expertise, though it reflects audience preference and may not fully capture all dimensions of performer skill. Additionally, editorial decisions may modify temporal structure: while editors likely preserve narrative coherence and pacing, we cannot rule out that editing systematically alters pause patterns. Future work can examine live performances to confirm whether these patterns persist without editorial mediation.

Second, our dataset focuses on Chinese stand-up comedy, raising generalizability questions. While temporal processing mechanisms are thought domain-general and culturally universal~\citep{large1999dynamics}, specific pause durations and timing-content relationships may vary across cultures and languages. Chinese is a tonal, character-based language, potentially affecting how timing cues semantic boundaries. Cross-linguistic replication would clarify which findings reflect universal cognitive mechanisms versus culturally specific norms.

Third, our semantic distance measures capture local sentence-to-sentence transitions but not higher-level narrative structure, callbacks, or thematic coherence. Comedy often involves long-range dependencies where jokes build on earlier setups or subvert expectations established minutes earlier. Future computational models should integrate hierarchical semantic representations to capture how local and global incongruity jointly shape experience, and may employ token surprisal to capture online prediction error at a finer grain.

Fourth, our analysis focused on features extractable from \ac{asr} transcripts, excluding prosodic features beyond timing (pitch, loudness, voice quality), gestural communication, audience interaction, and stage presence. The large effect sizes for pause-based features suggest timing alone captures substantial variance, but complete accounts require multimodal analysis integrating linguistic, paralinguistic, and visual cues.

Finally, the correlational design limits causal inference. We cannot determine whether temporal structure causally affects audience processing or reflects performer-side demands. Experimental studies manipulating pause durations while holding content constant are needed to establish causality.

Several directions could extend our framework. First, fine-grained audience response data (\eg, continuous laughter ratings, physiological measures) would reveal how timing-content coupling affects moment-by-moment engagement. Second, testing the \ac{dpv} framework across other humor formats (improvisation, sketch comedy) and related domains (music, rhetoric, storytelling) would clarify whether strategic coupling reflects humor-specific or domain-general principles. Third, examining individual differences would connect humor research with expertise and creativity: Do successful comedians consciously control timing-content coupling? Do audiences vary in sensitivity to temporal versus semantic violations? Addressing these questions would illuminate both humor cognition and predictive processing more generally.

\section{Conclusion}

By computationally analyzing 828 Chinese professional stand-ups, we establish that naturalistic humor success depends on coordinated violations of temporal and semantic predictions. Temporal control dominates overall effects, but strategic coupling---amplifying pauses before semantically surprising content---distinguishes professional expertise. These findings integrate humor theory with predictive processing frameworks, demonstrating that timing scaffolds semantic incongruity through cognitive gating mechanisms. Our \ac{dpv} framework advances understanding of how experts manipulate audience expectations in naturalistic settings, with implications for humor cognition, computational humor generation, and creativity research~\citep{ma2026narrativeloom,peng2025probing}.

\paragraph{Acknowledgement}

We thank Fengyuan Yang for creating elegant visuals. The work is supported by the National Natural Science Foundation of China (32595491, 62376009), the PKU-BingJi Joint Laboratory for Artificial Intelligence, the Wuhan Major Scientific and Technological Special Program (2025060902020304), the Hubei Embodied Intelligence Foundation Model Research and Development Program, and the National Comprehensive Experimental Base for Governance of Intelligent Society, Wuhan East Lake High-Tech Development Zone.

\printbibliography

\end{CJK*}
\end{document}

%% file: config.tex
\usepackage[utf8]{inputenc} 
\usepackage[T1]{fontenc}    
\usepackage{microtype}
\usepackage{times,latexsym}
\usepackage{graphicx} \graphicspath{{figures/}}
\usepackage{amsmath,amssymb,amsfonts,mathabx,mathtools,nicefrac}
\usepackage{acronym}
\usepackage[breaklinks,colorlinks,citecolor=black,bookmarks=true]{hyperref}
\usepackage{balance}
\usepackage{xspace}
\usepackage{setspace}
\usepackage[skip=3pt,font=small]{subcaption}
\usepackage[skip=3pt,font=small]{caption}
\usepackage[dvipsnames,svgnames,x11names,table]{xcolor}
\usepackage[capitalise,noabbrev,nameinlink]{cleveref}
\usepackage{booktabs,tabularx,colortbl,multirow,multicol,array,makecell,tabularray}
\usepackage{enumitem}
\usepackage[misc]{ifsym}
\usepackage{CJKutf8} 
\makeatletter
\DeclareRobustCommand\onedot{\futurelet\@let@token\@onedot}
\def\@onedot{\ifx\@let@token.\else.\null\fi\xspace}
\def\eg{\emph{e.g}\onedot}

\def\vs{\emph{vs}\onedot}

\makeatother

\makeatletter
\renewcommand{\paragraph}{%
    \@startsection{paragraph}{4}%
    {\z@}{0ex \@plus 0ex \@minus 0ex}{-1em}%
    {\normalfont\normalsize\bfseries}%
}
\makeatother

\usepackage[backend=biber,style=apa,natbib=true,annotation=false,backref=true]{biblatex}
\crefname{algorithm}{Alg.}{Algs.}
\Crefname{algocf}{Algorithm}{Algorithms}
\crefname{section}{Sec.}{Secs.}
\Crefname{section}{Section}{Sections}
\crefname{table}{Tab.}{Tabs.}
\Crefname{table}{Table}{Tables}
\crefname{figure}{Fig.}{Figs.}
\Crefname{figure}{Figure}{Figures}
\crefname{equation}{Eq.}{Eqs.}
\Crefname{equation}{Equation}{Equations}
\crefname{appendix}{Appx.}{Appxs.}
\Crefname{appendix}{Appendix}{Appendices}

\AtBeginDocument{
    \acrodef{dpv}[DPV]{Dual Prediction Violation}
    \acrodef{gtvh}[GTVH]{General Theory of Verbal Humor}
    \acrodef{asr}[ASR]{Automatic Speech Recognition}
}

%% file: reference.bib
@incollection{suls1972two,
  title={A two-stage model for the appreciation of jokes and cartoons: An information-processing analysis},
  author={Suls, Jerry},
  booktitle={The Psychology of Humor: Theoretical Perspectives and Empirical Issues},
  volume={1},
  pages={81--100},
  year={1972},
  publisher={Academic Press}
}

@incollection{suls1983cognitive,
  title={Cognitive processes in humor appreciation},
  author={Suls, Jerry},
  booktitle={Handbook of Humor Research: Volume 1: Basic Issues},
  pages={39--57},
  year={1983},
  publisher={Springer}
}

@article{attardo1991script,
  title={Script theory revis (it) ed: Joke similarity and joke representation model},
  author={Attardo, Salvatore and Raskin, Victor},
  journal={Humor: International Journal of Humor},
  volume={4},
  number={3},
  pages={293--347},
  year={1991}
}

@incollection{van2017affective,
  title={Affective value in the predictive mind},
  author={Van de Cruys, Sander and Metzinger, Thomas K and Wiese, W},
  booktitle={Philosophy and Predictive Processing},
  year={2017},
  pages={1--21},
  publisher={MIND Group; Frankfurt am Main}
}

@article{moran2004neural,
  title={Neural correlates of humor detection and appreciation},
  author={Moran, Joseph M and Wig, Gagan S and Adams Jr, Reginald B and Janata, Petr and Kelley, William M},
  journal={Neuroimage},
  volume={21},
  number={3},
  pages={1055--1060},
  year={2004},
  publisher={Elsevier}
}

@article{clark2018nice,
  title={A nice surprise? Predictive processing and the active pursuit of novelty},
  author={Clark, Andy},
  journal={Phenomenology and the Cognitive Sciences},
  volume={17},
  number={3},
  pages={521--534},
  year={2018},
  publisher={Springer}
}

@article{vrticka2013neural,
  title={The neural basis of humour processing},
  author={Vrticka, Pascal and Black, Jessica M and Reiss, Allan L},
  journal={Nature Reviews Neuroscience},
  volume={14},
  number={12},
  pages={860--868},
  year={2013},
  publisher={Nature Publishing Group}
}

@article{large1999dynamics,
  title={The dynamics of attending: how people track time-varying events.},
  author={Large, Edward W and Jones, Mari Riess},
  journal={Psychological Review},
  volume={106},
  number={1},
  pages={119},
  year={1999},
  publisher={American Psychological Association}
}

@article{jones1989dynamic,
  title={Dynamic attending and responses to time.},
  author={Jones, Mari Riess and Boltz, Marilyn},
  journal={Psychological Review},
  volume={96},
  number={3},
  pages={459},
  year={1989},
  publisher={American Psychological Association}
}

@article{barnes2000expectancy,
  title={Expectancy, attention, and time},
  author={Barnes, Ralph and Jones, Mari Riess},
  journal={Cognitive Psychology},
  volume={41},
  number={3},
  pages={254--311},
  year={2000},
  publisher={Elsevier}
}

@article{lehne2015toward,
  title={Toward a general psychological model of tension and suspense},
  author={Lehne, Moritz and Koelsch, Stefan},
  journal={Frontiers in Psychology},
  volume={6},
  pages={79},
  year={2015},
  publisher={Frontiers Media SA}
}

@article{lehne2015reading,
  title={Reading a suspenseful literary text activates brain areas related to social cognition and predictive inference},
  author={Lehne, Moritz and Engel, Philipp and Rohrmeier, Martin and Menninghaus, Winfried and Jacobs, Arthur M and Koelsch, Stefan},
  journal={PloS One},
  volume={10},
  number={5},
  pages={e0124550},
  year={2015},
  publisher={Public Library of Science San Francisco, CA USA}
}

@article{fiorillo2003discrete,
  title={Discrete coding of reward probability and uncertainty by dopamine neurons},
  author={Fiorillo, Christopher D and Tobler, Philippe N and Schultz, Wolfram},
  journal={Science},
  volume={299},
  number={5614},
  pages={1898--1902},
  year={2003},
  publisher={American Association for the Advancement of Science}
}

@article{bogels2011prosodic,
  title={Prosodic breaks in sentence processing investigated by event-related potentials},
  author={B{\"o}gels, Sara and Schriefers, Herbert and Vonk, Wietske and Chwilla, Dorothee J},
  journal={Language and Linguistics Compass},
  volume={5},
  number={7},
  pages={424--440},
  year={2011},
  publisher={Wiley Online Library}
}

@book{attardo2001humorous,
  title={Humorous texts: A semantic and pragmatic analysis},
  author={Attardo, Salvatore},
  volume={6},
  year={2001},
  publisher={Walter de Gruyter}
}

@article{norrick2001conversational,
  title={On the conversational performance of narrative jokes: Toward an account of timing.},
  author={Norrick, Neal R},
  journal={Humor: International Journal of Humor Research},
  volume={14},
  number={3},
  pages={255--274},
  year={2001}
}

@inproceedings{cattle2018recognizing,
  title={Recognizing humour using word associations and humour anchor extraction},
  author={Cattle, Andrew and Ma, Xiaojuan},
  booktitle={International Conference on Computational Linguistics},
  year={2018}
}

@inproceedings{mihalcea2010computational,
  title={Computational models for incongruity detection in humour},
  author={Mihalcea, Rada and Strapparava, Carlo and Pulman, Stephen},
  booktitle={International Conference on Intelligent Text Processing and Computational Linguistics},
  year={2010}
}

@article{veale2004incongruity,
  title={Incongruity in humor: Root cause or epiphenomenon?},
  author={Veale, Tony},
  journal={Humor: International Journal of Humor Research},
  volume={17},
  number={4},
  pages={419--428},
  year={2004}
}

@article{kao2016computational,
  title={A computational model of linguistic humor in puns},
  author={Kao, Justine T and Levy, Roger and Goodman, Noah D},
  journal={Cognitive Science},
  volume={40},
  number={5},
  pages={1270--1285},
  year={2016},
  publisher={Wiley Online Library}
}

@article{annamoradnejad2024colbert,
  title={ColBERT: Using BERT sentence embedding in parallel neural networks for computational humor},
  author={Annamoradnejad, Issa and Zoghi, Gohar},
  journal={Expert Systems with Applications},
  volume={249},
  pages={123685},
  year={2024},
  publisher={Elsevier}
}

@inproceedings{ma2025word,
  title={Word Embeddings Track Social Group Changes Across 70 Years in China},
  author={Ma, Yuxi and Peng, Yongqian and Zhu, Yixin},
  booktitle=CogSci,
  year={2025}
}

@inproceedings{ma2026narrativeloom,
  title={NarrativeLoom: Enhancing Creative Storytelling through Multi-Persona Collaborative Improvisation},
  author={Ma, Yuxi and Peng, Yongqian and Yang, Fengyuan and Zha, Siyu and Zhang, Chi and Jia, Zixia and Zheng, Zilong and Zhu, Yixin},
  booktitle=CHI,
  year={2026}
}

@inproceedings{peng2025probing,
  title={Probing and Inducing Combinational Creativity in Vision-Language Models},
  author={Peng, Yongqian and Ma, Yuxi and Wang, Mengmeng and Wang, Yuxuan and Wang, Yizhou and Zhang, Chi and Zhu, Yixin and Zheng, Zilong},
  booktitle=CHI,
  year={2025}
}


%% file: reference_header.bib
@string {CogSci = "{Annual Meeting of the Cognitive Science Society (CogSci)}"}

@string {CHI = "{ACM Conference on Human Factors in Computing Systems (CHI)}"}
